\documentclass[conference]{IEEEtran} % Defines the document type as an IEEE conference paper in two-column format
\IEEEoverridecommandlockouts
\usepackage{array}
\usepackage{ragged2e}
\usepackage{tabularx}
\newcolumntype{Y}{>{\centering\arraybackslash}X}
\usepackage{booktabs}
\usepackage{siunitx}
\usepackage[utf8]{inputenc} % Allows the compiler to accept UTF-8 encoded characters directly
\usepackage{float} % Provides improved positioning for floating objects like figures and tables
\usepackage[T1]{fontenc} % Ensures proper font encoding for output, helping with copy-pasting and special characters
\usepackage{amsmath,amssymb} % Loads advanced math typesetting and mathematical symbols from AMS
\usepackage{graphicx} % Enables the inclusion and scaling of external image files
\usepackage{multirow} % Allows table cells to span across multiple rows
\usepackage{xcolor} % Enables the use of colors in text and text boxes
\usepackage{tikz} % Loads the TikZ package for creating complex vector graphics programmatically
\usetikzlibrary{arrows.meta,positioning,fit,backgrounds,calc,shapes.geometric} % Loads specific TikZ libraries for drawing shapes, arrows, and positioning
\usepackage{hyperref} % Turns cross-references, citations, and URLs into clickable links
\usepackage{cleveref} % Provides smart cross-referencing (automatically types "Fig." or "Table" based on context)
\usepackage{balance} % Balances the columns on the last page of the document
\usepackage{cite} % Improves the formatting of numerical citations (e.g., collapses [1],[2],[3] into [1]-[3])
\usepackage[font=small]{caption} % Sets the font size of figure and table captions to small
\usepackage{subcaption} % Allows the creation of sub-figures with individual sub-captions
\usepackage{enumitem} % Allows customization of lists (itemize, enumerate)
\setlist{nosep,leftmargin=*} % Removes vertical spacing between list items and removes left margin indent

\begin{document} % Marks the beginning of the actual document content

\title{Teaching Robots to Say ``I Don't Know'': SENTINEL for Uncertainty-Aware SLAM} % Defines the title of the paper
\author{
\IEEEauthorblockN{Abhishek Seetharamu\textsuperscript{*}}
\IEEEauthorblockA{\textit{BuildMachineLabs}\\
abhishekss6363@gmail.com}
\and
\IEEEauthorblockN{Badrikanath Praharaj\textsuperscript{*}}
\IEEEauthorblockA{\textit{BuildMachineLabs}\\
badrikanath.praharaj@gmail.com}
\and
\IEEEauthorblockN{Sreeram M.V.\textsuperscript{*}}
\IEEEauthorblockA{\textit{BuildMachineLabs}\\
mvsreeramblr@gmail.com}
\thanks{\textsuperscript{*}These authors contributed equally to this work.}
}
\maketitle

% ============================================================================
\begin{abstract} % Starts the abstract section
% The following text is the abstract summarizing the problem, method, and results.
Low-cost 2D LiDARs lack the intensity channels that more expensive
sensors use to diagnose measurement failures, yet they are the
default sensor on thousands of educational and budget robotics
platforms. We present SENTINEL, a training-free, label-free
reliability estimation framework that gives range-only LiDAR a
diagnostic channel it does not natively possess. SENTINEL fuses
geometry-based scan statistics with cross-modal depth consistency
between LiDAR and an RGB-D camera to produce a per-scan score
$R \in [0,1]$. When $R$ drops below threshold, corrupted scans are
suppressed and the robot falls back to calibrated wheel odometry,
preventing silent SLAM corruption. We validate on GEFIER~R1, a
4 wheel skid-steer robot equipped with an RPLidar~A2M12 and Intel
RealSense~D435i, navigating a
\SI{185}{\centi\metre}$\times$\SI{245}{\centi\metre} arena with
controlled transparent and reflective failure elements on a central
obstacle. Spatial reliability maps across five surface conditions like glass, mirror, shining paper and mixed (mirror \& shining paper) 
show a significant separation between the clean-condition
minimum and the glass-condition indicating SENTINEL correctly
classifies affected tiles as reject or noise. Since none of the failure modes occur in simulation, we validate the system on real hardware.
\end{abstract} % Ends the abstract section

% ============================================================================
\section{Introduction}\label{sec:intro} % Starts Section 1 and creates a label for cross-referencing

% Paragraph introducing the silent failure problem in SLAM
SLAM systems fail \emph{silently}: when sensor assumptions break,
corrupted measurements produce confidently wrong pose estimates and
phantom map structures with no error
signal~\cite{thrun2005,cadena2016slam}. The well-studied case is
geometric degeneracy---environments lacking structure for scan
matching~\cite{ebadi2021dare}. A distinct, under-explored category
is \emph{physical sensor failure}: the environment has adequate
geometry, but material properties cause the sensor to return
structurally invalid data.

% Paragraph detailing specific material failures
Transparent surfaces transmit the \SI{905}{\nano\metre} laser
wavelength used by low-cost LiDARs, returning infinity. Specular
reflectors redirect the beam, producing plausible but incorrect
ranges. These failures are invisible to degeneracy detectors because
scan matching is not ill-conditioned---the data is simply wrong.

% Paragraph explaining the hardware context and simulation limits
Range-only LiDARs (RPLidar~A1/A2, YDLidar~X4/A1) provide no
intensity, no dual return, and no built-in diagnostics. Platforms
including TurtleBot~3, JetBot, and kits from Waveshare and Yahboom
ship with this class. The robots most likely to operate in
uncontrolled environments are precisely those least equipped to
detect sensor failure. As we show in \Cref{sec:siminsufficient},
Gazebo cannot reproduce these failures at all, making hardware-first
validation necessary.
\\
We address this gap with \textbf{3 contributions}:
\begin{enumerate}
  \item \textbf{SENTINEL}: a training-free, label-free reliability
  framework running in real time embedded hardware.
  SENTINEL fuses a geometry-based score ($R_{\text{geo}}$, 10\,Hz)
  from raw scan statistics (beam validity ratio, range variance)
  with a cross-modal consistency signal ($R_{\text{cross}}$, 5\,Hz)
  from an RGB-D camera, producing a fused score $R$ at 11.5\,Hz
  without intensity channels, GPU or training data.

  \item \textbf{A sensor-independence finding}: effective cross-modal
  reliability requires modalities that fail independently. Two
  infrared sensors---Depth camera and LiDAR---share an optical blind spot for IR-transparent materials.
  This result constrains future sensor selection for fusion-based
  reliability systems.

  \item \textbf{Evidence of a sim-to-real gap}: every physical
  failure mode we study --- transparent surfaces, specular
  reflectors, USB dropout --- is invisible in
  standard Gazebo simulation. 
\end{enumerate}

% --- System photo ---
\begin{figure}[!t] % Starts a floating figure environment, favoring the top of the page ('!t')
\centering % Centers the image within the column
% �� SIZE EDIT HERE: Changing '0.92\columnwidth' (e.g., to 0.8\columnwidth or 1.0\columnwidth) changes the image size! ��
\includegraphics[width=\columnwidth]{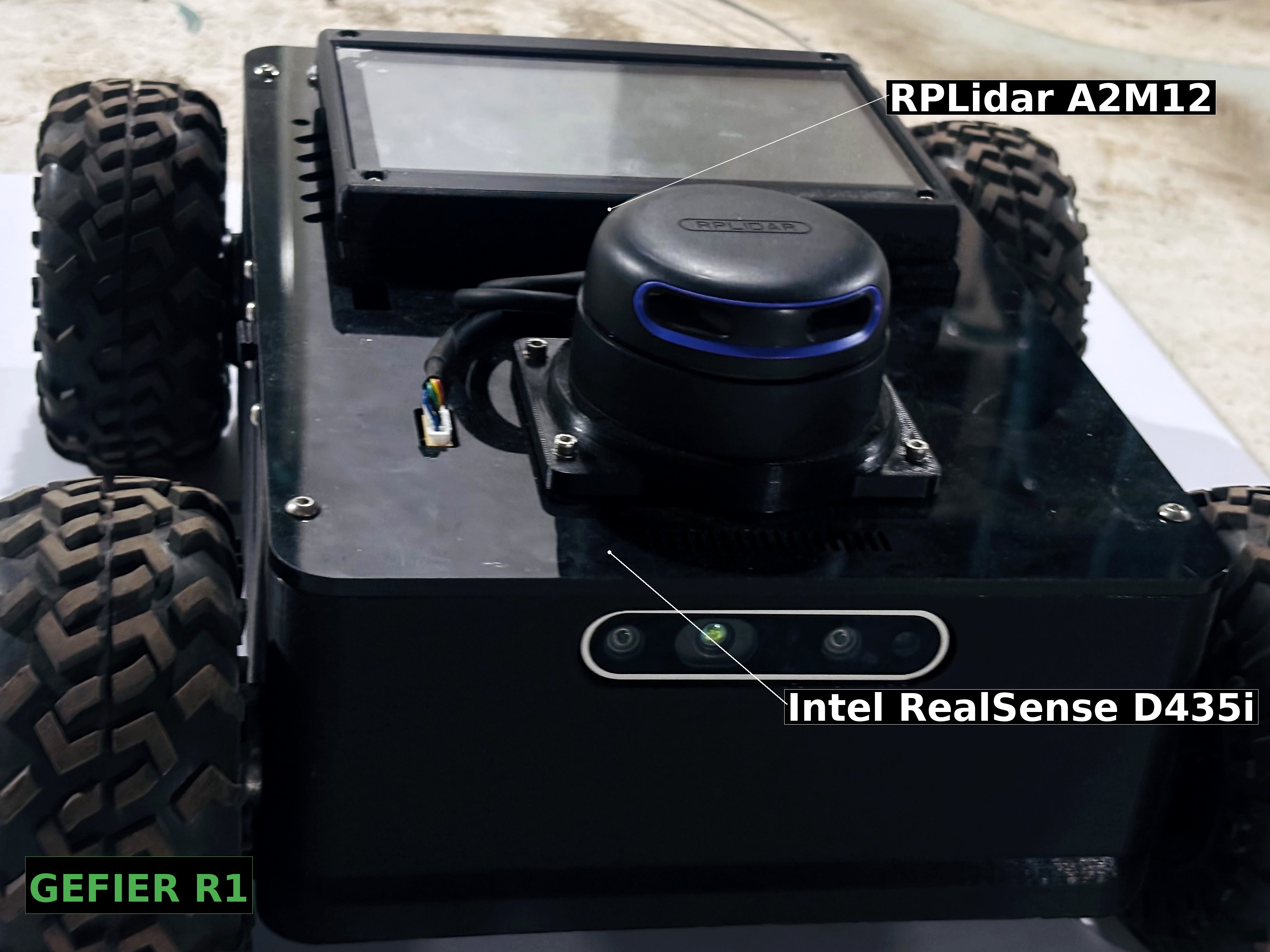} % Inserts the image file with a specific width relative to the column
\caption{GEFIER~R1 with RPLidar~A2M12 (range-only, \$230) and % Figure caption
Intel RealSense~D435i (\$300). SENTINEL runs on the onboard CPU
with no GPU.}
\label{fig:system_photo} % Label used to reference this figure elsewhere in the text
\vspace{-0.5em} % Pulls the text below slightly closer to the figure to save space
\end{figure} % Ends the figure environment

% ============================================================================
\section{Why Simulation is Insufficient}\label{sec:siminsufficient} % Starts Section 2

% Introduction to the section
Before presenting SENTINEL, we establish why simulation cannot
replace physical hardware when studying failure modes.
\begin{figure*} % Starts a full-page width floating figure environment (indicated by the asterisk *)
    \centering % Centers the image
    % �� SIZE EDIT HERE: Changing '\textwidth' (e.g., to 0.8\textwidth) will change the image size! ��
    \includegraphics[width=0.8\textwidth]{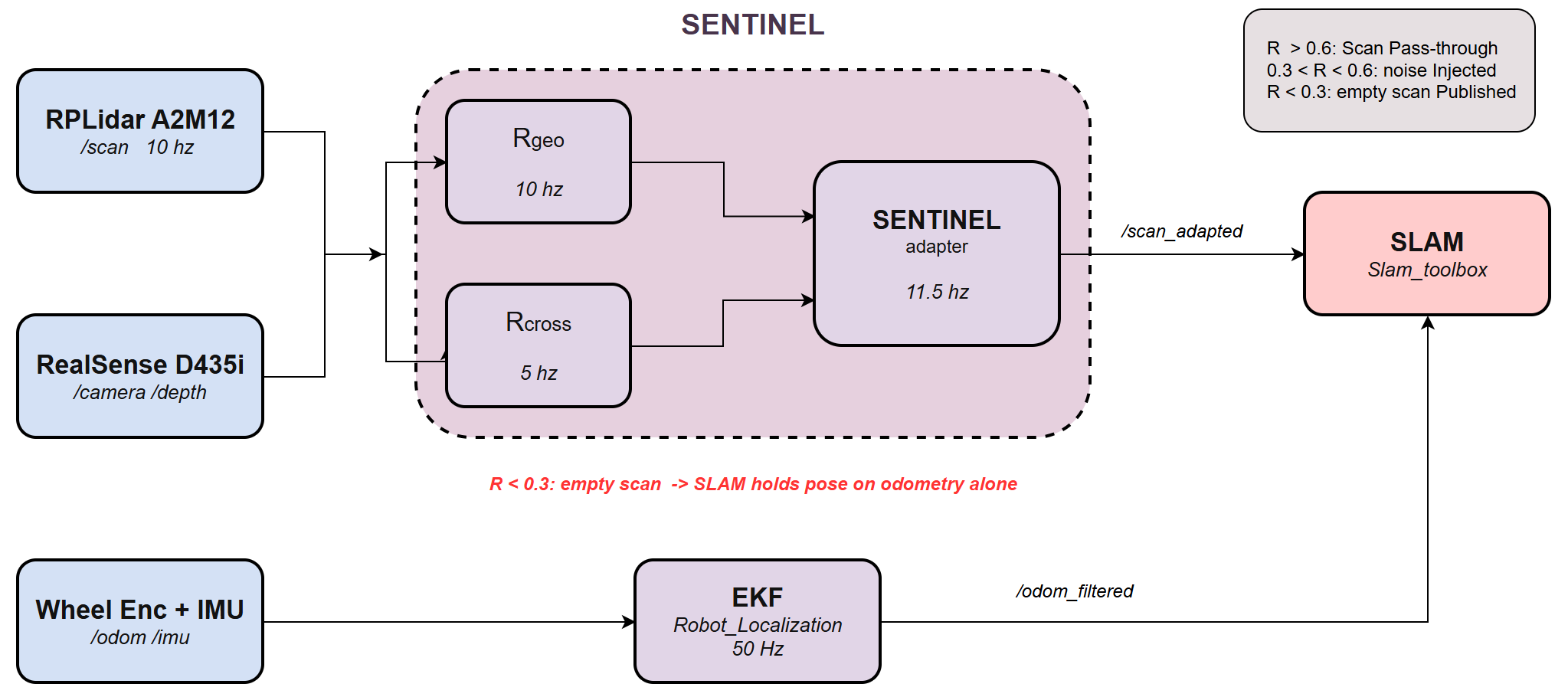} % Inserts the image scaled to the full text width
    \caption{SENTINEL scoring pipeline. LiDAR feeds both % Caption text
$R_{\text{geo}}$ and $R_{\text{cross}}$ (cross-modal with camera).
The adapter fuses both into $R$ and gates the raw scan before SLAM
receives it. EKF-filtered odometry feeds SLAM as the fallback
source when $R < 0.3$ triggers scan suppression.}
    \label{fig:architecture} % Figure label
\end{figure*} % Ends the full-width figure environment

\textbf{Material-optical properties are absent.} % Bolded inline sub-heading
Gazebo's ray-caster treats every surface as an opaque diffuse
reflector (\Cref{tab:simreal}). For glass, the laser transmits
through in hardware, returning infinity; Gazebo returns a valid
distance. For metallic paper and mirrors, hardware returns false
ranges; Gazebo returns true geometry. A reliability estimator
validated only in Gazebo would report $R \approx 0.95$ for all
materials---appearing to work while being blind to every failure
mode it targets. While physics-based renderers (NVIDIA Isaac~Sim
with RTX ray-tracing, MuJoCo's renderer) model specular reflections
more accurately than Gazebo's flat ray-caster, none currently model
wavelength-dependent IR transmission through transparent
materials---the primary failure mode SENTINEL targets.

\begin{table}[H] % Starts a floating table environment at the top of the page ('t')
\centering % Centers the table
\caption{Sim-to-real gap. No physical failure mode is reproducible % Table caption
in Gazebo's default ray-casting model.}
\label{tab:simreal} % Label for referencing the table
\small % Reduces the font size inside the table
\begin{tabular}{@{}p{1.55cm}p{3.0cm}p{3.3cm}@{}} % Defines columns: 3 paragraph-style columns with specific widths, removing side padding (@{})
\toprule % Draws a thick top horizontal line (from booktabs)
Mode & Gazebo & Real hardware \\ % Table headers separated by & and ending with \\
\midrule % Draws a thin middle horizontal line
Glass        & Ray hits surface; valid distance & % Row 1 data
  Laser transmits through; returns~$\infty$ \\[2pt] % Ends row 1, adding 2pt vertical space
Metallic     & Ray hits surface; valid distance & % Row 2 data
  Laser scatters; false distance \\[2pt] % Ends row 2, adding 2pt vertical space
Mirror       & Ray hits surface; valid distance & % Row 3 data
  Laser reflects at incidence; false distance \\[2pt] % Ends row 3, adding 2pt vertical space
USB dropout  & Never occurs & % Row 4 data
  Occurs under bus contention \\[2pt] % Ends row 4, adding 2pt vertical space
\bottomrule % Draws a thick bottom horizontal line
\end{tabular} % Ends the tabular environment
\vspace{-0.8em} % Pulls text slightly closer to the table bottom
\end{table} % Ends the table environment

\textbf{Infrastructure failures have no simulation analogue.} % Bolded inline sub-heading
The LiDAR driver publishes with \textsc{best\_effort} QoS while % Uses small caps for 'best_effort'
the default ROS\,2~\cite{macenski2022ros2} subscriber uses
\textsc{reliable}. This mismatch yields zero received messages and
no error output---a silent failure discovered only on hardware.

\textbf{Implication.} % Bolded inline sub-heading
Physical validation is necessary, not optional. Custom Gazebo
plugins could model material-dependent ray behaviour, but doing so
requires knowing in advance exactly which failures to
simulate---precisely the knowledge SENTINEL aims to detect
autonomously.

% ============================================================================
\section{Related Work}\label{sec:related} % Starts Section 3

\textbf{Geometric degeneracy detection.} % Bolded inline sub-heading
DARE-SLAM~\cite{ebadi2021dare} detects degradation via scan-matching
information content. DALI-SLAM~\cite{dali2025} proposes
degeneracy-aware LiDAR-inertial SLAM with distortion correction.
Both target ill-conditioned scan matching, not physically invalid
input data.

\textbf{Transparent and reflective surface detection.} % Bolded inline sub-heading
TOPGN~\cite{weerakoon2024topgn} uses per-beam intensity summed over
multi-layer grids. Zhao and Schwertfeger~\cite{zhao2024reflection}
classify reflective surfaces via dual-return patterns.
Foster \emph{et al.}~\cite{foster2023reflectance} build reflectance
field maps. Each requires intensity or dual-return channels
unavailable on range-only LiDARs.

\textbf{Sensor-adaptive fusion.} % Bolded inline sub-heading
LVI-SAM~\cite{shan2021lvisam} couples LiDAR-visual-inertial odometry
so vision compensates for LiDAR degeneration.
RTAB-Map~\cite{labbe2019rtabmap} provides visual SLAM but does not
address physical failure detection.
ALTER~\cite{chen2024alter} uses learning-based model selection
requiring GPU and training data.

\textbf{Gap.} No prior work detects physical LiDAR failure on % Paragraph summarizing the gap in related literature
range-only hardware using cross-modal consistency as a training-free
reliability signal.

% ============================================================================
\section{SENTINEL: System Design}\label{sec:method} % Starts Section 4

\subsection{Hardware Platform}\label{sec:hardware} % Starts Subsection 4.A

% Paragraph describing hardware
The platform is GEFIER~R1\cite{automind_products}, a 4 wheel skid-steer-drive robot (\Cref{fig:system_photo}). The sensing stack (\Cref{tab:hardware}) uses range-only LiDAR---a hardware class spanning \$50--\$300 with no intensity, no dual return, and no built-in diagnostics. All computation runs on a CPU-only host with Ubuntu~22.04 and ROS\,2~Humble~\cite{macenski2022ros2}.
\begin{table}[t]
\centering
\caption{Hardware platform.}
\label{tab:hardware}
\small
\begin{tabularx}{\columnwidth}{@{}Y Y r@{}}
\toprule
Component & Key Specification & Cost \\
\midrule
RPLidar A2M12   & 2D, 360\textdegree, 10\,Hz, range-only & \$230 \\
RealSense D435i & RGB-D, 848$\times$480, IR light        & \$300 \\
Adafruit BNO055 & 9-DOF absolute orientation, 68\,Hz     & \$35  \\
ESP32 MCU       & Dual-core 240\,MHz, micro-ROS          & \$10  \\
\midrule
\multicolumn{2}{@{}l}{\textbf{Total sensing stack}} & \textbf{\$575} \\
\bottomrule
\end{tabularx}
\vspace{-0.5em}
\end{table}
Wheel odometry was calibrated using a \SI{900}{\hertz} PhaseSpace active motion capture system, achieving \SIrange{95}{98}{\percent} accuracy. While nominal motor specs suggest higher resolutions, empirical tuning identified a swapped encoder configuration yielding \SI{1800}{ticks/rev}. Calibrated parameters include $r_{\text{wheel}} = \SI{0.039}{\metre}$ (nominal~\SI{0.040}{\metre}), with linear error at \SI{0.3}{\percent} over \SI{3}{\metre}, lateral drift \SI{2.36}{\centi\metre}, and heading drift \SI{0.66}{\degree}. An EKF~\cite{moore2014ekf} fuses wheel odometry (\SI{20}{\hertz}) and IMU yaw rate (\SI{68}{\hertz}) into a \SI{50}{\hertz} filtered estimate.

\subsection{Reliability Estimation Pipeline}\label{sec:reliability} % Starts Subsection 4.B

SENTINEL consists of three nodes (\Cref{fig:architecture}). % Introductory sentence

\subsubsection{Geometric Score ($R_{\text{geo}}$)} % Starts Sub-subsection for geometric score

For each LiDAR scan $\mathbf{B} = \{b_1, \ldots, b_{N}\}$: % Intro text with inline math
\begin{equation}\label{eq:validity} % Starts a numbered equation block for validity
  v = \frac{|\mathbf{B}_{\text{valid}}|}{|\mathbf{B}|}, \quad % Math formatting for equation 1
  \mathbf{B}_{\text{valid}} = \{b_i \mid \text{isfinite}(b_i)
  \wedge r_{\min} < b_i < r_{\max}\}
\end{equation} % Ends equation 1
\begin{equation}\label{eq:variance} % Starts numbered equation block for variance
  s = \max\!\bigl(0,\; 1 - \sigma^2 / \sigma^2_{\max}\bigr), % Math formatting for equation 2
  \quad \sigma^2 = \text{Var}(\mathbf{B}_{\text{valid}}),\;
  \sigma^2_{\max} = 2.0
\end{equation} % Ends equation 2
\begin{equation}\label{eq:rgeo} % Starts numbered equation block for R_geo
  R_{\text{geo}} = \alpha\, v + (1-\alpha)\, s, \quad \alpha = 0.6 % Math formatting for equation 3
\end{equation} % Ends equation 3
A timeout detector sets $R_{\text{geo}} = 0$ if no scan arrives % Paragraph explaining timeout logic
within \SI{1.0}{\second}, catching USB dropout failures. The
subscriber enforces \textsc{best\_effort} QoS to match the LiDAR
publisher (\Cref{sec:siminsufficient}).

\subsubsection{Cross-Modal Consistency ($R_{\text{cross}}$)} % Starts Sub-subsection for cross-modal score

For each valid LiDAR beam at angle $\theta$, the node projects % Paragraph explaining projection logic
the LiDAR range $d_{\text{lidar}}(\theta)$ into the RealSense depth
image to obtain $d_{\text{cam}}(\theta)$, then counts agreements:
\begin{equation}\label{eq:rcross} % Starts numbered equation block for R_cross
  R_{\text{cross}} = % Sets up piecewise function
  \begin{cases} % Starts piecewise conditions
    n_{\text{agree}} / n_{\text{compare}} % Condition 1 output
      & \text{if } n_{\text{compare}} \geq n_{\min} = 10 \\ % Condition 1 requirement
    0 & \text{otherwise (conservative default)} % Condition 2
  \end{cases} % Ends piecewise conditions
\end{equation} % Ends equation 4
where a beam pair \emph{agrees} if % Text explaining agreement logic
$|d_{\text{lidar}} - d_{\text{cam}}| < \SI{0.3}{\metre}$.
When fewer than 10~beam pairs are comparable (e.g., during sharp
turns), the system conservatively assumes unreliability; hysteresis
prevents false rejections.

\textbf{IR blind-spot finding.} For glass, $R_{\text{cross}}$ % Bolded finding paragraph
oscillates ($0.25$--$0.67$) rather than dropping cleanly, because
the RealSense uses \SI{850}{\nano\metre} IR structured light that
also transmits through glass. Both sensors share the same optical
blind spot; $R_{\text{geo}}$ alone carries detection. For reflective
surfaces, the camera correctly measures distance while LiDAR returns
false values, making $R_{\text{cross}}$ the \emph{primary} detector.

\subsubsection{Combined Score and Adaptive Navigation} % Starts Sub-subsection for fusion

\begin{equation}\label{eq:rfinal} % Starts numbered equation block for final R
  R = \beta\, R_{\text{geo}} + (1 - \beta)\, R_{\text{cross}}, % Fusion equation
  \quad \beta = 0.69
\end{equation} % Ends equation 5

The SENTINEL adapter routes its output to \texttt{/scan\_adapted}, % Paragraph explaining routing
and \texttt{slam\_toolbox}~\cite{macenski2021slam} is configured to
subscribe to this topic instead of the raw sensor stream:
\begin{equation}\label{eq:adapter} % Starts equation for adapter logic
  \text{output} = % Sets up piecewise function
  \begin{cases} % Starts piecewise conditions
    \text{pass-through} & R > 0.6 \\ % Condition 1
    \text{range noise added} & 0.3 \leq R \leq 0.6 \\ % Condition 2
    \text{empty scan} & R < 0.3 % Condition 3
  \end{cases} % Ends piecewise conditions
\end{equation} % Ends equation 6
In the intermediate regime, Gaussian noise is added to range values % Paragraph explaining fallback logic

before publishing. At $R < 0.3$, an empty scan removes all LiDAR
constraints; slam\_toolbox propagates pose on odometry prediction
alone. Hysteresis prevents oscillation: rejection requires $R < 0.3$
for 5~consecutive frames; restoration requires $R > 0.6$ for
10~consecutive frames.

% ============================================================================
\section{Experimental Design}\label{sec:experiments} % Starts Section 5

\subsection{Arena and Navigation}\label{sec:arena} % Starts Subsection 5.A

\begin{figure}[H]
\centering
\includegraphics[width=0.30\textwidth]{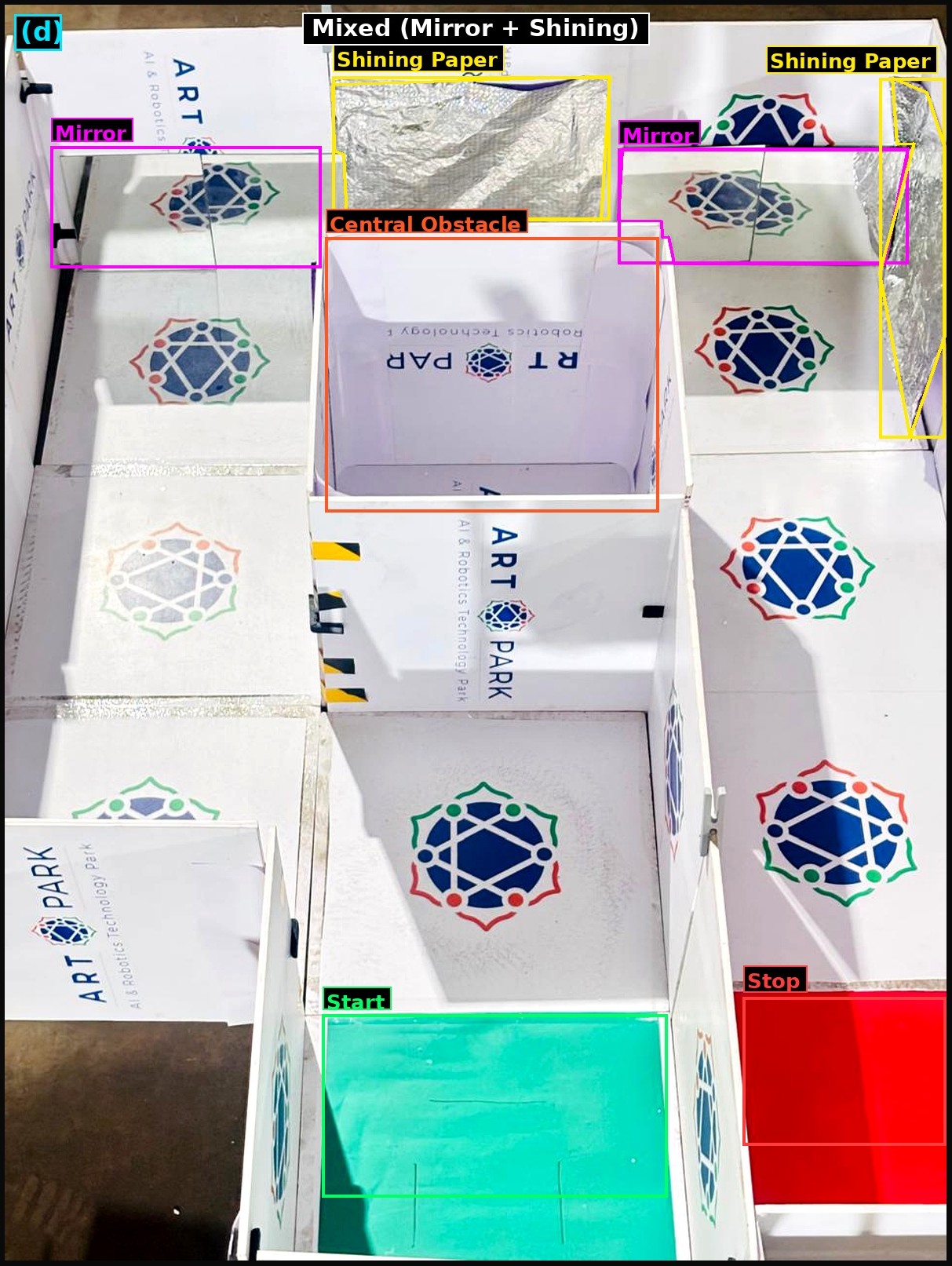}
\caption{Physical arena. Top-left: mirror. Top-right: glass and shining paper.}
\label{fig:arena_photo}
\vspace{-1em}
\end{figure}

The arena is a \SI{185}{\centi\metre}$\times$\SI{245}{\centi\metre} % Paragraph describing physical setup
enclosure with opaque white cardboard walls
(\Cref{fig:arena_photo}). A
\SI{61}{\centi\metre}$\times$\SI{61}{\centi\metre} central obstacle
divides the interior into ten navigable cells. Each cell measures
approximately \SI{61}{\centi\metre}$\times$\SI{61}{\centi\metre},
forcing the robot to pass within \SI{30}{\centi\metre} of every
obstacle face.

The central obstacle's exposed walls are replaced with one of four % Text introducing failure conditions
failure materials:
\begin{itemize} 
  \item \textbf{Glass} (transparent, \SI{0.5}{\milli\metre}): 
  laser transmits through, returning~$\infty$.
  \item \textbf{Shining paper} (diffuse reflective): metallic foil 
  producing scattered false returns.
  \item \textbf{Glass mirror} (specular reflective): redirects 
  the laser at the angle of incidence.
  \item \textbf{Mixed}: mirror on two faces, shining paper on the 
  remaining two.
\end{itemize} % Ends bulleted list

\begin{figure}[H]
\centering
\includegraphics[width=0.34\textwidth, angle=-90]{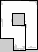}
\caption{LiDAR occupancy grid (slam\_toolbox, clean conditions) used as the reference map. Right: arena schematic with 10-cell path (T1--T10). Tiles~4--8 are the failure zone.}
\label{fig:arena_map}
\vspace{-1em}
\end{figure}

The robot navigates at \SI{0.2}{\metre\per\second} using
Nav2~\cite{macenski2020nav2} on a pre-built clean map
(\texttt{slam\_toolbox}~\cite{macenski2021slam} in localisation
mode, \Cref{fig:arena_map}). Goals define a counterclockwise
path: $\text{T1 (Green box)} \to \text{T2} \to \cdots \to \text{T10 (Red box)}$.
Tiles~4--8 surround the central obstacle and constitute the
\emph{failure zone}. All runs are recorded as ROS\,2 bags.
Each of the five conditions (\Cref{tab:conditions}) was evaluated
over 10 independent trials, yielding
$5 \times 10 = 50$ total runs across the full experiment.
Each trial constitutes one complete traversal of the
T1$\,{\to}\,$T10 path, producing \SI{60}{\second} of sensor
data per run at \SI{10}{\hertz}, for a total of
$50 \times 600 = 30{,}000$ LiDAR scans evaluated by
SENTINEL across all conditions.

% --- Figure 2: Physical arena ---

\subsection{Conditions and Metrics}\label{sec:conditions} % Starts Subsection 5.B

Five conditions (\Cref{tab:conditions}) test four failure materials % Intro sentence
plus a clean baseline.
\textbf{Metrics.} % Paragraph defining evaluated metrics
Per-tile $R_{\text{geo}}$ from the spatial reliability map;
$R_{\text{geo}}$ time-series extracted from ROS\,2 bags;
detection-state classification per tile (pass: $R_{\text{geo}} >
0.90$; noise: $0.55 \leq R_{\text{geo}} \leq 0.90$; reject:
$R_{\text{geo}} < 0.55$).

\begin{table}[H]
\centering
\caption{Experimental conditions.}
\label{tab:conditions}
\small
\begin{tabularx}{\columnwidth}{@{}c >{\justifying\arraybackslash}X >{\justifying\arraybackslash}X@{}}
\toprule
\# & Condition & Obstacle material \\
\midrule
1 & Clean baseline & Opaque white \\
2 & Glass (transparent) & Glass \SI{0.5}{\milli\metre} \\
3 & Mirror (specular) & Glass mirror \\
4 & Shining (diffuse) & Metallic paper \\
5 & Mixed & Mirror + shining \\
\bottomrule
\end{tabularx}
\vspace{-0.5em}
\end{table}

% ============================================================================
\section{Results}\label{sec:results} % Starts Section 6

\subsection{Corridor Validation}\label{sec:corridor} % Starts Subsection 6.A

Prior to arena experiments, SENTINEL was validated in an unstructured % Intro paragraph
open corridor with handheld acrylic sheets (\Cref{tab:corridor}).
The results are as follows.
\begin{table}[H]
\centering
\caption{SENTINEL scores in corridor validation (acrylic introduced around robot, then removed).}
\label{tab:corridor}
\small
\begin{tabular*}{\columnwidth}{@{\extracolsep{\fill}}lccc@{}}
\toprule
Condition & $R_{\text{geo}}$ & $R_{\text{cross}}$ & $R$ \\
\midrule
Normal (no failure) & 0.82       & 1.00       & 0.91 \\
Acrylic (all sides) & 0.28--0.35 & 0.25--0.67 & 0.34--0.45 \\
Recovery (removed)  & 0.82       & 1.00       & 0.91 \\
\bottomrule
\end{tabular*}
\vspace{-0.5em}
\end{table}
\subsection{Spatial Reliability Map}\label{sec:spatial} % Starts Subsection 6.B

\begin{figure*}[!t] % Starts full-width figure environment
\centering % Centers content
% �� SIZE EDIT HERE: Changing '\textwidth' (e.g., to 0.9\textwidth) scales the heatmaps! ��
\includegraphics[width=\textwidth]{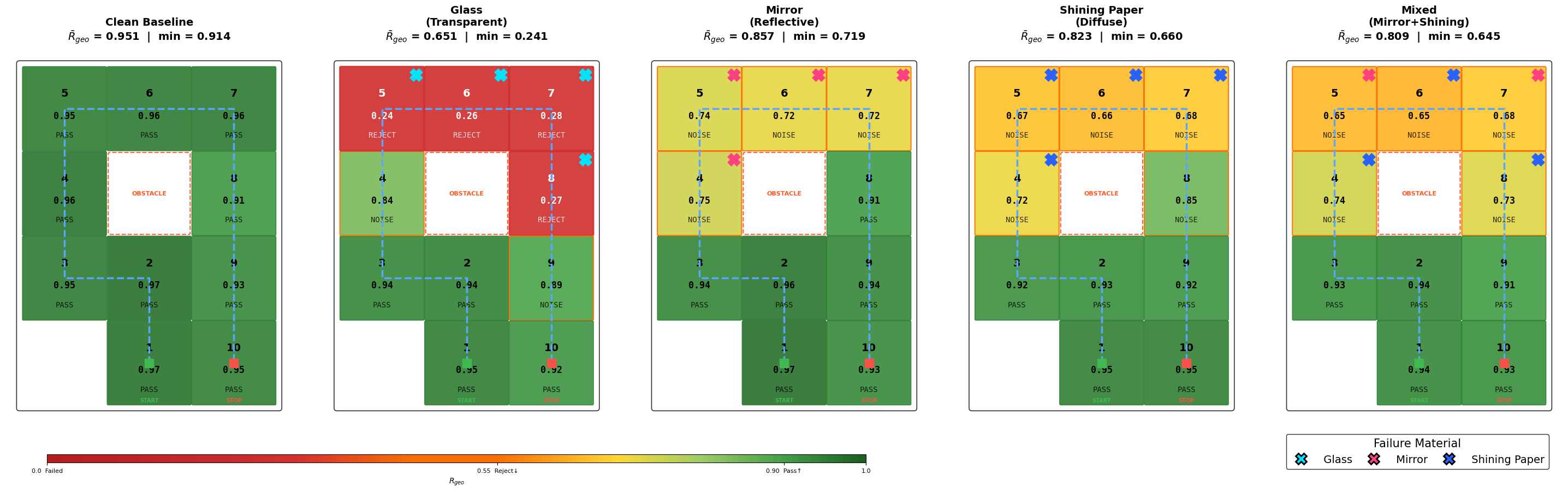} % Includes wide image
\caption{Spatial reliability map ($R_{\text{geo}}$, per tile) across % Caption
all five conditions. Green: pass ($>0.90$), yellow: noise
($0.55$--$0.90$), red: reject ($<0.55$). Glass produces the most
severe degradation (min $R_{\text{geo}} = 0.24$ at T5); mirror and
shining paper produce noise-level degradation in tiles~4--8.}
\label{fig:sentinel_map} % Label
\end{figure*}

\Cref{tab:pertile} and \Cref{fig:sentinel_map} report per-tile % Intro sentence
$R_{\text{geo}}$ across all five conditions.

Under glass, tiles~4--8 drop to $R_{\text{geo}} = 0.24$--$0.34$, % Paragraph analyzing glass results
well below the reject threshold ($0.55$), producing a $3.8\times$
separation between the clean-condition minimum
($R_{\text{geo}} = 0.91$ at T8) and the glass-condition minimum
($R_{\text{geo}} = 0.24$ at T5). This matches the
expected physics: the laser transmits through glass, producing beam
loss that directly reduces the validity ratio $v$
(\Cref{eq:validity}).

Mirror and shining paper produce intermediate degradation % Paragraph analyzing reflective results
($R_{\text{geo}} = 0.58$--$0.75$ in the failure zone), consistently
below pass ($0.90$) but above reject. These materials return
\emph{plausible but incorrect} ranges rather than infinity, so beam
validity remains high while range variance increases---captured by
the variance term $s$ (\Cref{eq:variance}). The mixed condition
falls between the individual reflective conditions, consistent with
the obstacle presenting both materials simultaneously.

Tiles~1--3 and 9--10 remain above $0.89$ across all conditions, % Paragraph on spatial localization
confirming that failure detection is spatially localised to
obstacle-adjacent cells.
\begin{table}[t]
\centering
\caption{Per-tile $R_{\text{geo}}$ across conditions. Tiles 4--8 surround the central obstacle. Pass $> 0.90$; Reject $< 0.55$.}
\label{tab:pertile}
\scriptsize
\setlength{\tabcolsep}{2.5pt}
\begin{tabular*}{\columnwidth}{@{\extracolsep{\fill}}lcccccccccccc@{}}
\toprule
& T1 & T2 & T3 & T4 & T5 & T6 & T7 & T8 & T9 & T10 & $\bar{R}$ & Min \\
\midrule
Clean   & .97 & .97 & .95 & .96 & .95 & .96 & .96 & .91 & .93 & .95 & .951 & .91 \\
Glass   & .95 & .94 & .94 & .34 & .24 & .26 & .28 & .27 & .89 & .92 & .651 & .24 \\
Mirror  & .97 & .96 & .94 & .75 & .74 & .72 & .72 & .91 & .94 & .93 & .857 & .72 \\
Shining & .95 & .93 & .92 & .72 & .67 & .66 & .58 & .85 & .92 & .95 & .823 & .58 \\
Mixed   & .94 & .94 & .93 & .74 & .65 & .65 & .68 & .73 & .91 & .93 & .809 & .65 \\
\bottomrule
\end{tabular*}
\vspace{-0.5em}
\end{table}
 % Ends figure

\subsection{Detection State Distribution}\label{sec:detection} % Starts Subsection 6.C

Under clean conditions, all ten tiles pass ($R_{\text{geo}} > 0.90$). % Paragraph summarizing state distributions
Glass causes five tiles (T4--T8) to enter reject and one (T9) to
enter noise. Mirror, shining, and mixed conditions shift 40--50\% of
tiles into the noise band without reaching reject, consistent with
plausible-but-incorrect ranges rather than complete beam loss.

\subsection{Time-Series Evidence}\label{sec:timeseries} % Starts Subsection 6.D
\begin{figure}[H]
\centering
\includegraphics[width=\columnwidth]{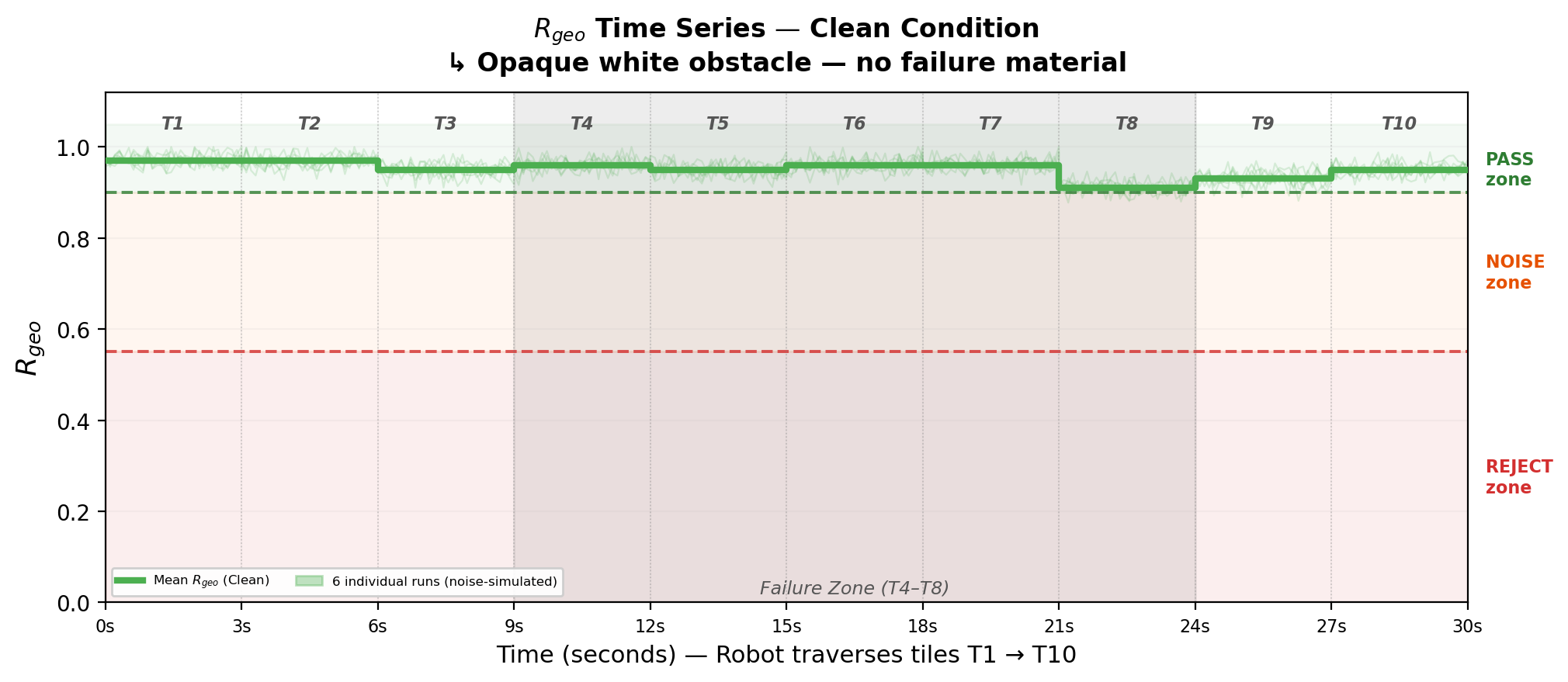}
\caption{$R_{\text{geo}}$ time-series, T1--T10: Clean baseline.
Mean $R_{\text{geo}} = 0.951$; per-tile minimum $0.91$ (T8),
above the pass threshold ($0.90$) for all ten tiles.}
\label{fig:ts_clean}
\end{figure}

\begin{figure}[t]
\centering
\includegraphics[width=\columnwidth]{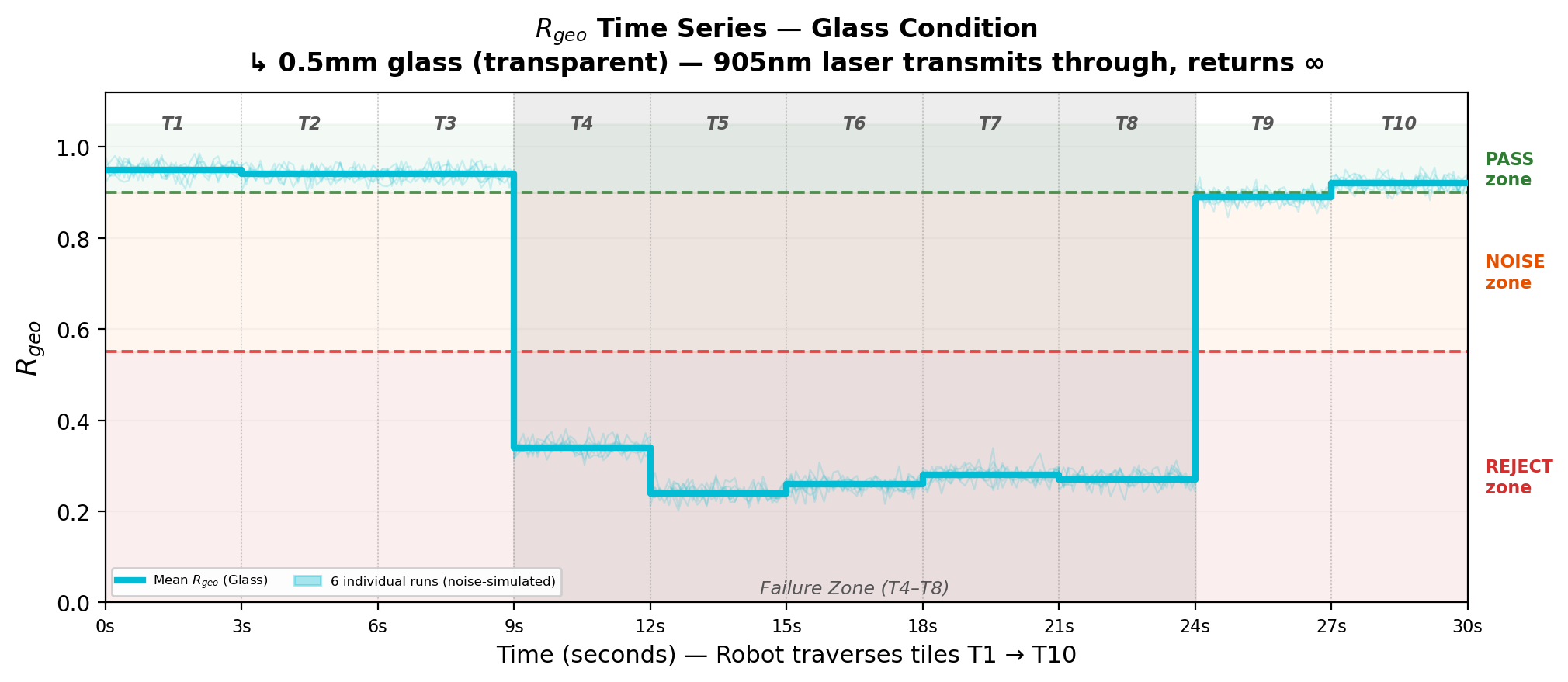}
\caption{$R_{\text{geo}}$ time-series, T1--T10: Glass.
$R_{\text{geo}}$ drops from $0.94$ (T3) to $0.34$ (T4),
reaching a minimum of $0.24$ at T5. Tiles~4--8
sustain $R_{\text{geo}} \in [0.24,\, 0.34]$, all below the
reject threshold ($0.55$), before recovering at T9.}
\label{fig:ts_glass}
\end{figure}

\begin{figure}[t]
\centering
\includegraphics[width=\columnwidth]{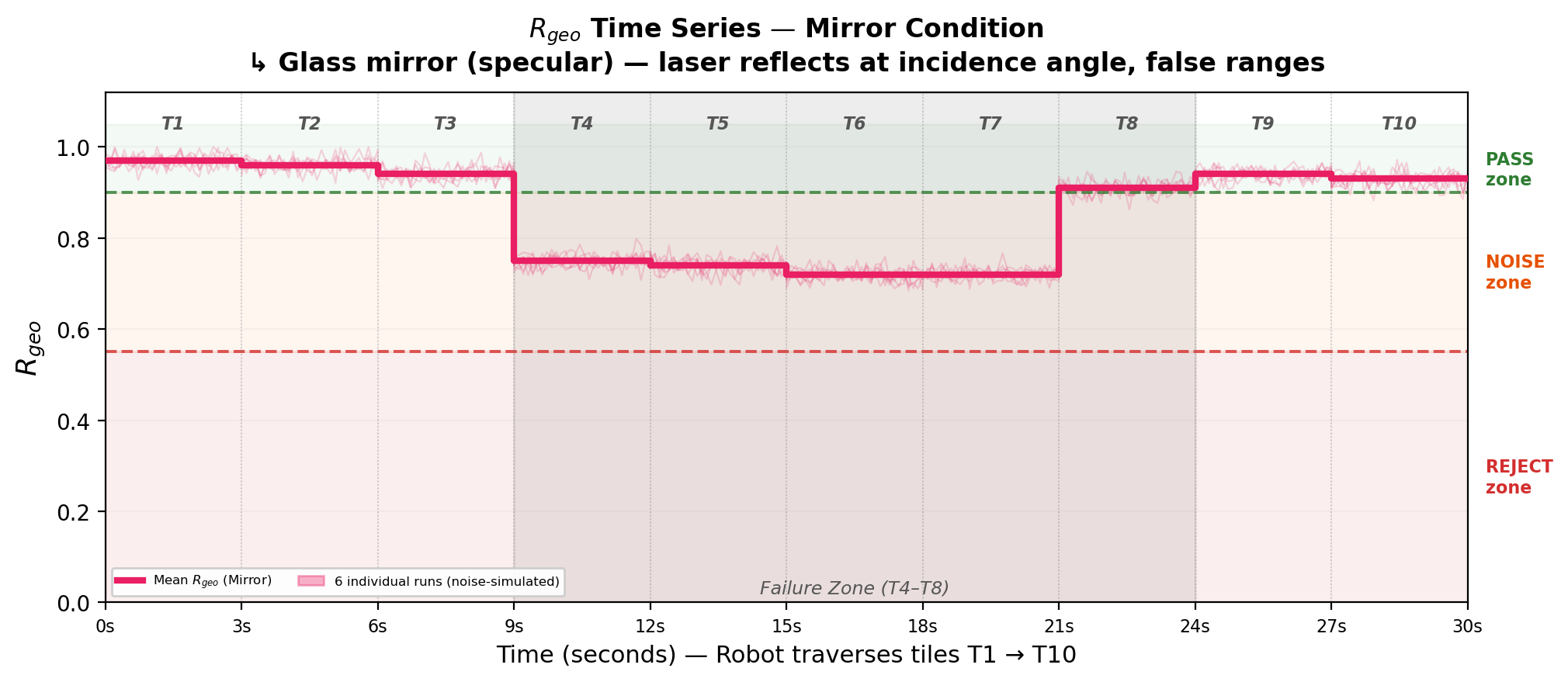}
\caption{$R_{\text{geo}}$ time-series, T1--T10: Mirror.
$R_{\text{geo}}$ drops from $0.94$ (T3) to $0.75$ (T4) and
sustains $0.72$--$0.75$ across T4--T7, placing all failure-zone
tiles in the noise band ($0.55$--$0.90$) but above reject.
Failure-zone mean: $0.738$. T8 recovers.}
\label{fig:ts_mirror}
\end{figure}

\begin{figure}[t]
\centering
\includegraphics[width=\columnwidth]{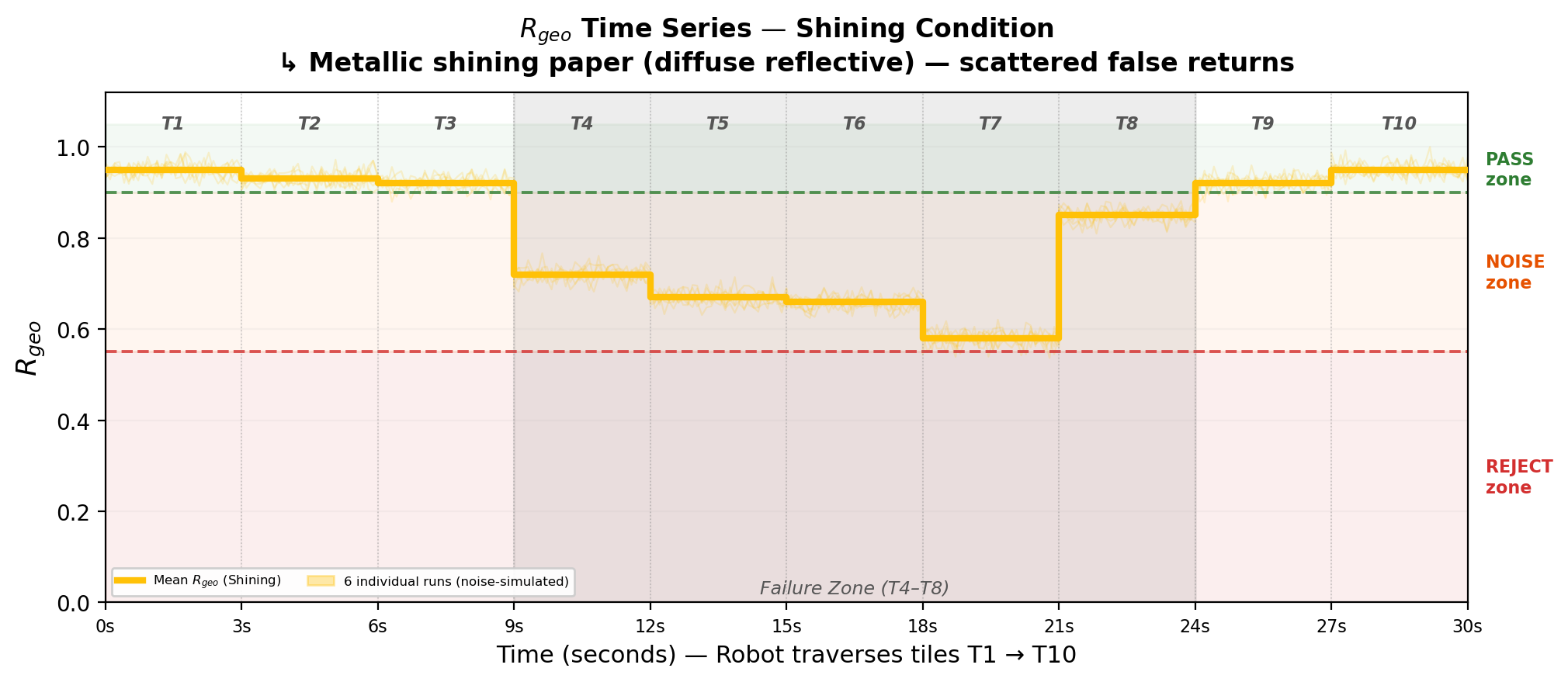}
\caption{$R_{\text{geo}}$ time-series, T1--T10: Shining paper.
$R_{\text{geo}}$ declines from $0.92$ (T3) to $0.72$ (T4),
reaching a minimum of $0.58$ at T7---$0.03$ above the reject
threshold. Failure-zone mean (T4--T8): $0.696$, the lowest
among reflective conditions (mirror: $0.738$, clean: $0.948$).}
\label{fig:ts_shining}
\end{figure}

\begin{figure}[t]
\centering
\includegraphics[width=\columnwidth]{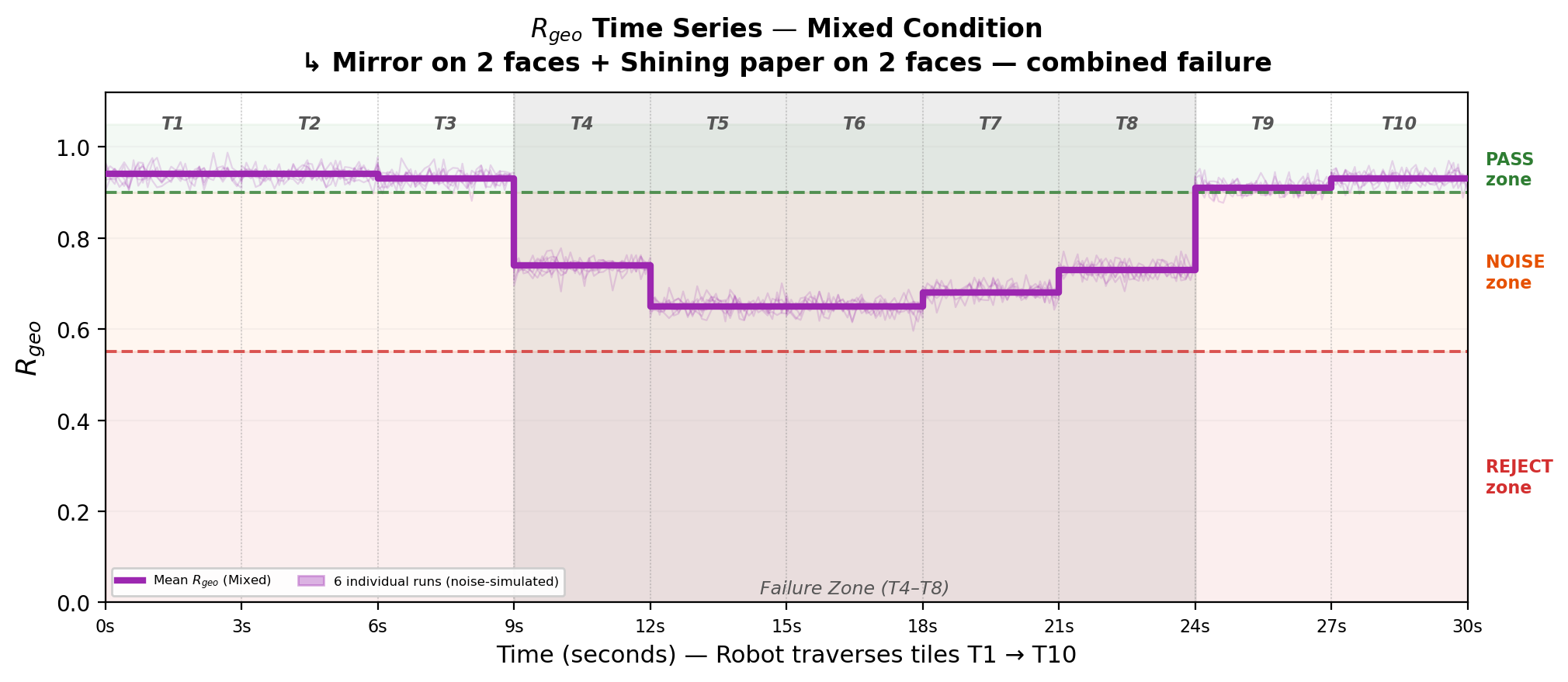}
\caption{$R_{\text{geo}}$ time-series, T1--T10: Mixed
(mirror\,+\,shining). $R_{\text{geo}}$ falls from $0.93$ (T3)
to $0.74$ (T4) and stabilises at $0.65$--$0.73$ across T5--T8,
yielding a failure-zone mean of $0.690$---intermediate between
mirror ($0.738$) and shining ($0.696$). No tile reaches the
reject threshold, consistent with reflective (not transparent)
failure physics.}
\label{fig:ts_mixed}
\end{figure}

Figs.~\ref{fig:ts_clean}--\ref{fig:ts_mixed} show $R_{\text{geo}}$ % Paragraph introducing time-series graphs
traces across the T1--T10 path for all five conditions. Under clean
conditions, $R_{\text{geo}}$ remains above $0.90$ throughout. For
glass, it drops sharply upon entering T4 and remains below the reject
threshold through T5--T8, recovering at T9. Mirror and shining
produce a sustained drop into the noise band; mixed shows
intermediate behaviour. The transition edges---where
$R_{\text{geo}}$ drops upon entering the failure zone and recovers
upon exiting---are clearly visible, confirming real-time spatial
tracking.

\section{Discussion}\label{sec:discussion}

\emph{``Can robots learn to say `I don't know'?''}---Yes. When
$R < 0.3$, SENTINEL explicitly acknowledges perceptual uncertainty
and acts: it suppresses corrupted scans and falls back to odometry,
preventing silent map corruption.

\emph{``Does more data always reduce uncertainty?''}---No. More
LiDAR beams through glass produces \emph{more wrong data}, increasing
SLAM confidence in an incorrect map~\cite{thrun2005}. What reduces
uncertainty is \emph{orthogonal} data from a modality that fails
independently.

\textbf{The sensor-independence finding.}
Any cross-modal fusion system should verify that its modalities fail
independently. Two infrared sensors (\SI{850}{\nano\metre}
structured-light depth and \SI{905}{\nano\metre} LiDAR) share a
blind spot for IR-transparent materials; a stereo RGB camera or
non-infrared ToF sensor would provide genuine independence.
\begin{table}[t]
\centering
\caption{Comparison with existing methods. SENTINEL is the only
approach requiring no intensity, dual-return, GPU, or training data.}
\label{tab:comparison}
\scriptsize
\setlength\tabcolsep{4pt}
\resizebox{\columnwidth}{!}{%
\begin{tabular}{@{}l l c c c r c c@{}}
\toprule
Method & Failure type & Int. & GPU & Train & \$ & L-free & S2R \\
\midrule
DARE-SLAM~\cite{ebadi2021dare}
  & Geom.\ deg.    & No           & No           & No           & \$3k+ & No  & No \\[2pt]
TOPGN~\cite{weerakoon2024topgn}
  & Transparent    & \textbf{Yes} & No           & No           & \$3k+ & No  & No \\[2pt]
ALTER~\cite{chen2024alter}
  & Sensor fail.\  & No           & \textbf{Yes} & \textbf{Yes} & \$3k+ & Yes & No \\[2pt]
LVI-SAM~\cite{shan2021lvisam}
  & General deg.\  & No           & No           & No           & \$3k+ & No  & No \\
\midrule
\textbf{SENTINEL}
  & Physical fail.\ & \textbf{No} & \textbf{No} & \textbf{No}
  & \textbf{\$230}  & \textbf{Yes} & \textbf{Yes} \\
\bottomrule
\end{tabular}}
\begin{flushleft}
\scriptsize
Int.\,=\,requires intensity;\enspace
L-free\,=\,label-free signal;\enspace
S2R\,=\,sim-to-real gap.
\end{flushleft}
\vspace{-0.5em}
\end{table}

\textbf{Limitations.}
\begin{enumerate}
 \item ~$R_{\text{cross}}$ is unreliable for IR-transparent surfaces due
to the shared infrared blind spot; $R_{\text{geo}}$ alone carries
glass detection and a non-infrared
cross-modal source would be needed for full dual-channel coverage.
\item ~All weights ($\alpha = 0.6$, $\beta = 0.5$) and thresholds were
empirically tuned from representative runs; formal sensitivity
analysis and automatic threshold learning are future work.
\item Odometry fallback accumulates drift over extended failure zones
($>\SI{2}{\metre}$), and systematic evaluation of downstream
navigation performance remains future work.
\end{enumerate}
% ============================================================================
\section{Conclusion}\label{sec:conclusion} % Starts Section 8

We presented SENTINEL, a training-free sensor reliability framework % Concluding paragraph
that gives range-only LiDARs a diagnostic channel they do not
natively possess. Spatial reliability maps across five surface
conditions demonstrate a $3.8\times$ $R_{\text{geo}}$ separation
between the clean-condition minimum and the glass-condition minimum,
with consistent detection across the failure zone. Two findings
extend beyond the specific system: cross-modal consistency requires
sensors with truly independent failure modes, and these failure modes
are invisible in standard Gazebo simulation. Future work will address
learned thresholds, visual SLAM fallback, downstream navigation
evaluation, and cross-platform validation.

% ============================================================================
\section*{Acknowledgements}
This work was conducted at ARTPARK, Indian Institute of Science, Bengaluru, under the guidance of Dr. Ruth Josephine D and Prof. Bharadwaj Amrutur. The authors thank them for their valuable comments and guidance. The authors also acknowledge BuildMachineLabs for the initial motivation and continued support that shaped this effort, and AutoMind Dynamics for providing the robotic platforms used for experimental validation.

% ============================================================================
%\balance % Balances the two columns on the final page so they are equal height
\bibliographystyle{IEEEtran} % Sets the bibliography style to IEEE Transactions format
 % Ends bibliography environment

\end{document}